\documentclass[letterpaper, 10 pt, conference]{ieeeconf}  

\IEEEoverridecommandlockouts                              
\usepackage{verbatim}
\usepackage{mathtools}
\usepackage{amssymb}
\usepackage{enumerate}
\usepackage{tikz}
\usepackage{tikzscale}
\usepackage[hidelinks]{hyperref}
\usepackage[multiple]{footmisc}
\usepackage{graphicx}
\usepackage{array}
\usepackage{gensymb}
\usepackage{cite}
\usetikzlibrary{positioning}
\usetikzlibrary{shapes.geometric, arrows}
\let\savedegree\degree
\let\degree\relax
\usepackage{mathabx}
\let\degree\savedegree

\newcommand{\blue}[1]{\textcolor{blue}{#1}}

\newcommand{\updrive}{\textit{\mbox{UP-Drive} }}
\newcommand{\nclt}{\textit{NCLT }}
\newcommand{\kitti}{\textit{KITTI }}

\newcommand{\ncltns}{\textit{NCLT}}
\newcommand{\kittins}{\textit{KITTI}}
\newcommand{\maptracking}{\mbox{map-tracking} }

\newcommand{\gloloc}{global localization }

\newcommand{\cf}{\mathcal{F}}
\newcommand{\bo}{B}
\newcommand{\ma}{M}

\newcommand{\bodyframe}{\cf_{\bo}}
\newcommand{\mapframe}{\cf_{\ma}}

\newcommand{\poseab}[2]{T^{#1}_{#2}}
\newcommand{\poset}[1]{\poseab{#1}{\ma \bo}}
\newcommand{\pose}{T_{\ma \bo}}
\newcommand{\poseguess}[1]{\hat{T}^{#1}_{\ma \bo}}

\newcommand{\state}{\mathbf{x}}

\newcommand{\odometry}{\bar{T}^{odo}_{{\bo_{\timestep} \bo_{\timestep+1}}}}
\newcommand{\fprior}{f_{prior}}
\newcommand{\fodo}{f_{odo}}
\newcommand{\floc}{f_{loc}}
\newcommand{\cov}[1]{P^{-1}_{#1}}
\newcommand{\covodo}{Q}
\newcommand{\cost}{c}

\newcommand{\ddt}{\delta}
\newcommand{\rt}{\rho}
\newcommand{\recallmt}{\mathbf{r}_{mt}}
\newcommand{\recalllc}{\mathbf{r}_{gl}}
\newcommand{\transmedianerror}{\mathbf{\bar{p}^{e}_{xyz}}}
\newcommand{\transerror}{\mathbf{p^{e}_{xyz}}}
\newcommand{\planarmedianerror}{\mathbf{\bar{p}^{e}_{xy}}}
\newcommand{\planarerror}{\mathbf{p^{e}_{xy}}}
\newcommand{\latmedianerror}{\mathbf{\bar{p}^{e}_{y}}}
\newcommand{\laterror}{\mathbf{p^{e}_{y}}}
\newcommand{\orientmedianerror}{\mathbf{\bar{\theta}^{e}_{xyz}}}
\newcommand{\orienterror}{\mathbf{\theta^{e}_{xyz}}}
\newcommand{\avginliers}{\mathbf{obs}_{\diameter}}

\newcommand{\timestep}{t}

\newcommand{\visloc}{VIZARD }
\newcommand{\vislocns}{VIZARD}
\newcommand{\sevensensesymbol}{2}
\newcommand{\aslsymbol}{1}

\tikzstyle{startstop} = [rectangle, rounded corners, minimum width=3cm, minimum height=1cm,text centered, draw=black, fill=red!30]
\tikzstyle{io} = [trapezium, trapezium left angle=70, trapezium right angle=110, minimum width=0cm, minimum height=0cm, text centered, draw=black, fill=blue!30]
\tikzstyle{process} = [rectangle, minimum width=0cm, minimum height=0cm, text centered, text width=3cm, draw=black, fill=orange!30]
\tikzstyle{decision} = [diamond, minimum width=0cm, minimum height=0cm, text centered, draw=black, fill=green!30]
\tikzstyle{arrow} = [thick,->,>=stealth]

\title{\LARGE \bf
\vislocns: Reliable Visual Localization for Autonomous Vehicles in Urban Outdoor Environments
}

\author{Mathias B\"{u}rki$^{\aslsymbol,\sevensensesymbol}$, Lukas Schaupp$^{\aslsymbol}$, Marcin Dymczyk$^{\sevensensesymbol}$, Renaud Dub\'e$^{\sevensensesymbol}$, Cesar Cadena$^{\aslsymbol}$,\\ Roland Siegwart$^{\aslsymbol}$, and Juan Nieto$^{\aslsymbol}$
\\ \small$^{\aslsymbol}$Autonomous Systems Lab, ETH Z\"{u}rich, {\tt\footnotesize \{firstname.lastname\}@mavt.ethz.ch}
\\
$^{\sevensensesymbol}$Sevensense Robotics AG, {\tt\footnotesize \{firstname.lastname\}@sevensense.ch}
}

\begin{document}

\maketitle
\thispagestyle{empty}
\pagestyle{empty}

\begin{abstract}
Changes in appearance is one of the main sources of failure in visual localization systems in outdoor environments.
To address this challenge, we present \vislocns, a visual localization system for urban outdoor environments.
By combining a local localization algorithm with the use of multi-session maps, a high localization recall can be achieved across vastly different appearance conditions.
%
%
The fusion of the visual localization constraints with wheel-odometry in a state estimation framework further guarantees smooth and accurate pose estimates.
In an extensive experimental evaluation on several hundreds of driving kilometers in challenging urban outdoor environments, we analyze the recall and accuracy of our localization system, investigate its key parameters and boundary conditions, and compare different types of feature descriptors. 
Our results show that \visloc is able to achieve nearly $100\%$ recall with a localization accuracy below $0.5m$ under varying outdoor appearance conditions, including at night-time.
\end{abstract}

\section{Introduction}
Localization is a pivotal capability of any autonomous vehicle.
%
By knowing their precise location, vehicles are able to plan a path to a next goal location, navigate safely in the environment, and eventually successfully complete their mission.
Especially for autonomous vehicles in urban environments, localization is challenging, as GNSS based localization systems fail to provide reliable and precise enough localization near buildings due to multi-path effects, or in tunnels or parking garages due to a lack of visible satellites.
Alternative exteroceptive sensor modalities are therefore necessary to accomplish this task, of which LiDARs and cameras have received most attention in recent years~\cite{cadena2016past, Lowry2016}.
While LiDARs have become more suited for mass market adoption, we believe there are still significant advantages with camera-based localization systems, despite the challenges related to long-term appearance change in outdoor environments.
Cameras remain considerably more cost-effective than LiDAR sensors, allowing them to be deployed in multitudes and in a flexible way on a large quantity of vehicles.
Furthermore, they can be used for sensing both appearance and geometric information of the environment, and are often better suited for global localization and loop-closure detection, which are necessary capabilities for bootstrapping any local localization algorithm, and to maintain geometrically consistent maps in lifelong operation~\cite{cadena2016past}.

\begin{figure}
\includegraphics[width=0.5\textwidth]{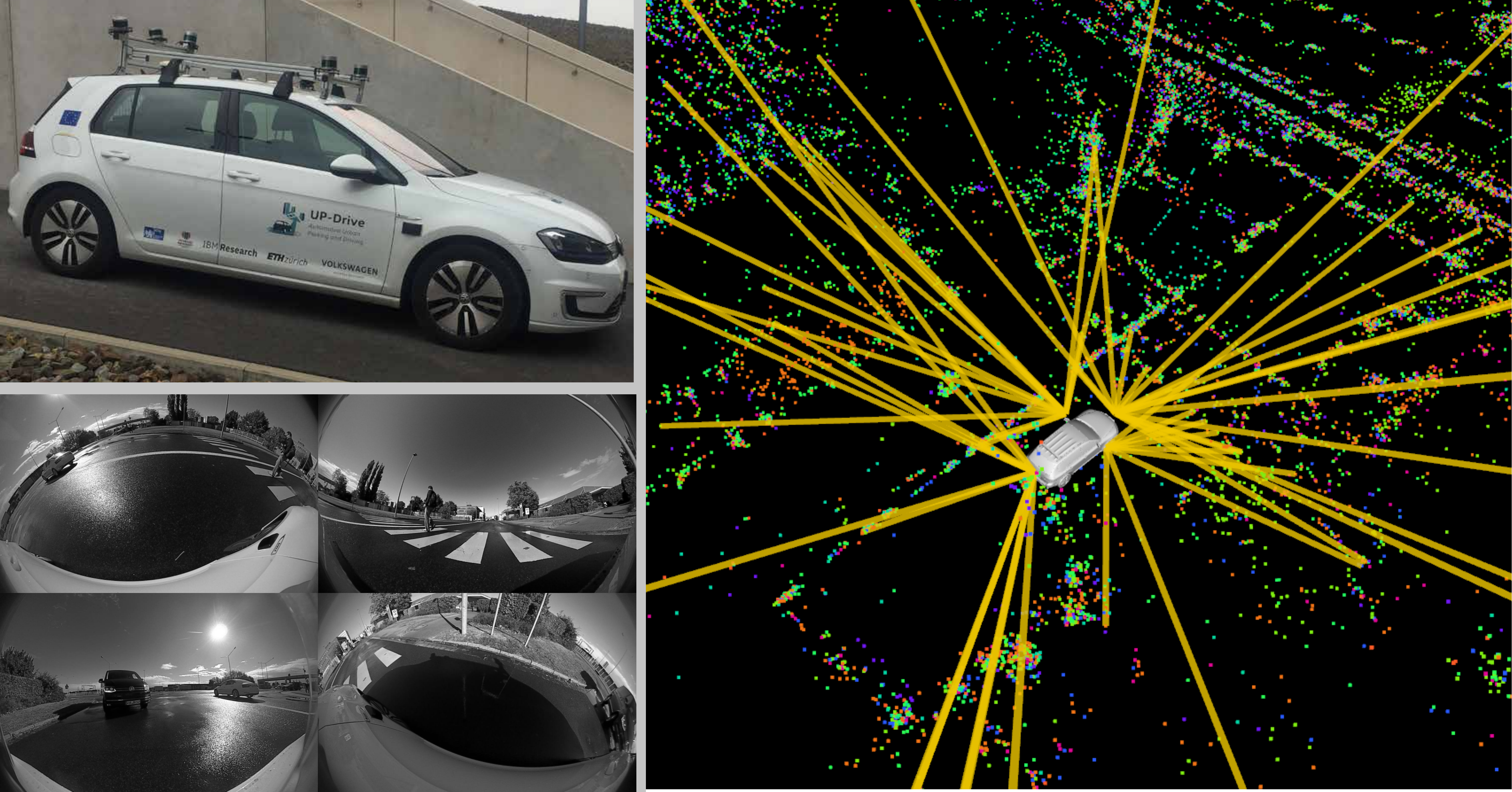}
\caption{\label{fig:teaser}
We aim at accurately localizing the \updrive vehicle depicted in the upper-left corner in a map of visual features depicted on the right side.
Features are extracted from images of the surround-view camera system (lower-left corner) and matched against $3D$ landmarks in the map.
Inlier matches, centered on the estimated $6DoF$ pose of the vehicle in the map, are illustrated as dark yellow lines on the right side. 
%
}
\vspace{-4mm}
\end{figure}
For these reasons, we have developed a visual localization system, dubbed \vislocns, for the self-driving cars in the \updrive project\footnote{The \textbf{UP-Drive} project is a research endeavor funded by the European Commission, aiming at advancing research and development towards fully autonomous cars in urban environment. See \url{www.up-drive.eu}.}, with the following main features:
\begin{enumerate}
\item We employ map-tracking, a local localization algorithm able to generate both accurate $6DoF$ pose estimates and achieve high localization recall.
\item Multi-session mapping techniques enable us to successfully tackle the challenge of long-term appearance change in outdoor environments, and even allow for localizing in night-time conditions.
\item The use of binary descriptors and an efficient sensor fusion backend renders real-time localization with CPU-only hardware set-ups feasible.
\end{enumerate}
In a thorough evaluation of all crucial aspects of our localization system using two long-term outdoor dataset collections, one of them publicly available, we carefully analyze the most important parameters in our pipeline,
compare the use of different binary descriptors, investigate key performance metrics such as localization accuracy and recall and relate to a state-of-the-art metric global localization algorithm.
We see the main added value of this paper in sharing with the community the insights gained in this long-term study.
%
%
%
%
%

The contributions of this paper are thus as follows:
\begin{itemize}
\item We thoroughly study the critical parameters of our localization system, analyze their boundary conditions, and share our gained insights.
\item From a comparison of the localization performance using different binary descriptors, we show which descriptors are best suited for map-tracking across long-term appearance change.
\item In an extensive evaluation on multiple long-term dataset collections, we demonstrate state-of-the-art localization performance across vastly different appearance conditions in outdoor environments, including at night time.
%
%
\end{itemize}

\section{related work}
Visual localization systems can be divided into two main categories. 
Global localization systems are able to retrieve the location of a robot with no prior knowledge of the robot's pose.
In contrast to that, local localization systems exploit a motion model to compute a prior on the robot location, thereby reducing the search space in the map.

\subsubsection*{Global Localization}
\label{sec:related_work:subsec:global_localization}
Early visual global localization systems have been presented in the context of offline geometric scene reconstruction from a large number of images collected from varying viewpoints~\cite{Schindler2007CityScale, Agarwal2009}.
These works led the foundation for many subsequent $6DoF$ global localization algorithms, and have been improved in numerous follow-up works~\cite{Li2010LocationRecogition, Sattler2011FastImage, liu2017efficient, Lynen2014PlacelessPlace, Cummins2011AppearanceOnlySLAM}.
More recently, deep learning techniques have given rise to novel global localization algorithms with remarkable robustness against drastic appearance change~\cite{detone2017superpoint, sarlin2018coarse}.
They require, however, high-end GPUs in order to achieve real-time operation.

In general, the aforementioned global localization algorithms are capable of achieving reliable localization across significant appearance change in outdoor environments. 
However, as shown in \cite{sattler2018benchmarking}, they often fall short of providing high recall with localization accuracies below $0.5m$, and are thus not well suited for our application, where we aim at permanently localizing our vehicle with sufficient accuracy to prevent deviation from the road lane boundaries.
%
Note that there has also been a substantial amount of work on global localization in the realm of place recognition, or image retrieval \cite{milford2012seqslam, naseer2014robust, maddern2012cat, torii201524, arandjelovic2016netvlad}.
These approaches, however, only provide a best matching image candidate in a map, instead of a $6DoF$ metric pose, and are thus addressing a different problem than ours.

\subsubsection*{Local Localization}
\label{sec:related_work:subsec:local_localization}
Local localization algorithms take prior information on the robot pose into account, in order to reduce the localization search space and increase recall. 
This is well motivated in practice, as subsequent localization attempts of a mobile robot are far from independent, but in fact highly correlated in space, with the incremental motion between images often observable, although with drift, from odometry sensors such as wheel-odometry or IMUs.
As a consequence, instead of regarding localization as an independent module, it can be directly integrated into the state estimation framework that optimizes the robot's pose in its environment by fusing odometry measurements and localization constraints.
The ORB-SLAM~\cite{mur2017visual} framework with its \textit{localization mode} offers a local localization system similar to ours.
%
%
They lack, however, the capability to integrate multiple sessions into a map, which greatly limits the robustness towards appearance change in outdoor environments.
Lategahn and Schiller present a hierarchical visual localization system for outdoor environments that combines a global with a local localization module~\cite{Lategahn2014}.
They achieve robustness against appearance change by employing DIRD descriptors \cite{Lategahn2014DIRD}. 
Their experimental evaluation, however, only spans across six weeks, and it thus remains unclear, how well their system performs over long-term appearance change.
Instead of employing an illumination invariant descriptor, Paton et al. use color-constant images to gain robustness against appearance change~\cite{Paton2015ItsNot}.
However, color-constancy primarily removes shadows under sunlight, but does not tackle other variations in appearance, such as seasonal change, or transitions from day to night-time.

\subsubsection*{Multi-Session Mapping}
\label{sec:related_work:subsec:ms_mapping}
A common technique to achieve robustness against arbitrary long-term appearance change incorporates visual cues from multiple sorties through the environment in the map.
We refer to this as multi-session mapping.
Schneider et al. have presented a state estimation framework that fuses visual-inertial odometry with metric global localization~\cite{schneider2018maplab,bloesch2017iterated,lynen2015get}. 
While their mapping framework allows merging several sessions, they use a feature based global localization algorithm which prohibits sufficient localization recall in outdoor environments.
Paton et al. present a visual localization system using multi-session mapping in~\cite{Paton2016Bridging}.
%
Their application is, however, restricted to a teach-and-repeat scenario.
In contrast to that, we employ loop-closure detection and bundle adjustment in order to get geometrically consistent multi-session maps, which adds additional flexibility in route planning and navigation.
The ``Experienced-Based Mapping'' framework developed by Churchill et al. maintains separate map instances for diverse appearance conditions~\cite{Churchill2013ExperienceBasedNavigation}.
This allows visual localization in arbitrarily appearance conditions in an elegant and efficient manner.
However, the maintenance of separate maps for differing conditions renders it impossible to share visual cues between sessions, which can increase recall.
Furthermore, an integration of the localization module into a complete navigation stack is more challenging, as the visual pose estimates are expressed with respect to separate, disconnected coordinate frames.

Similarly to our localization system,
the works presented in~\cite{Muhlfellner2016SummaryMaps,Muehlfellner2013Evaluation,lauer2017mapping,sons2018efficient} use a local localization algorithm with multi-session maps for localization.
As opposed to our work, M\"uhlfellner et al., refrain from fusing their visual pose estimates with wheel-odometry, which limits the accuracy and smoothness of their pose estimation framework, while Sons et al. do not report on localization accuracy and recall in long-term experiments.
\section{Methodology}
\label{sec:methodology}
\begin{figure}
\includegraphics[width=0.49\textwidth]{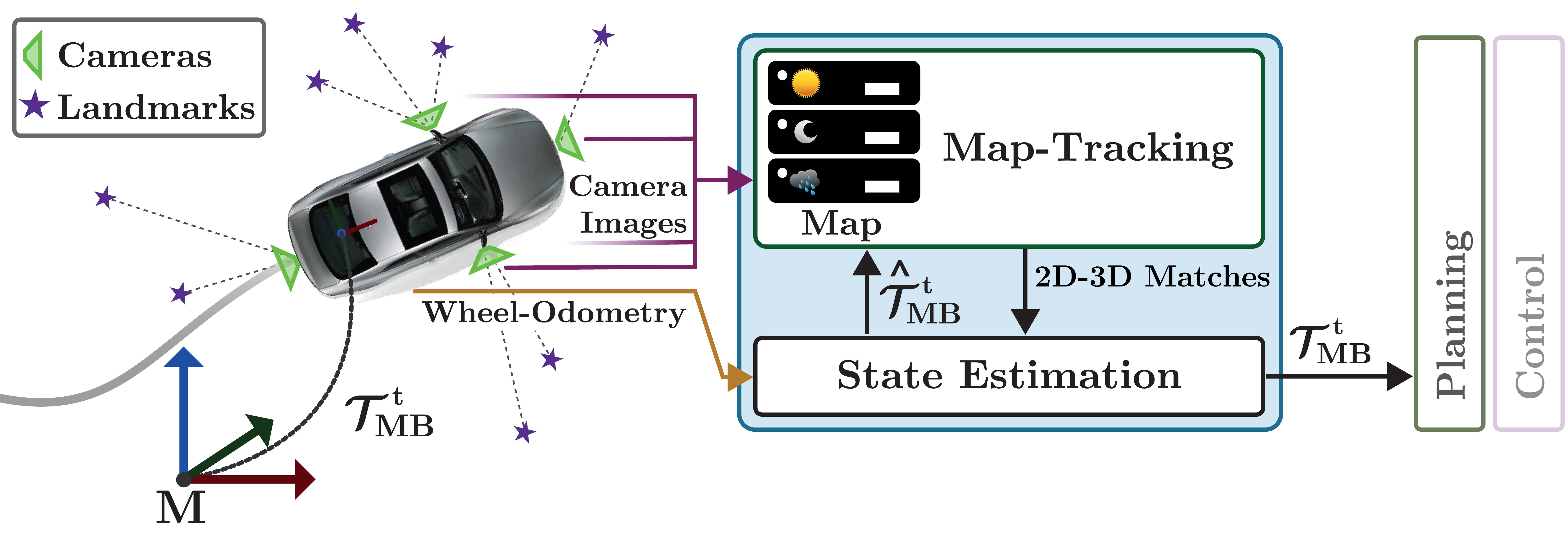}
\caption{\label{fig:schema}
The map-tracking module extracts 2D features from current camera images, and matches them with 3D map landmarks locally in image space using a pose prior $\poseguess{\timestep}$.
The state estimation module fuses the visual \mbox{2D-3D} matches with the current wheel-odometry measurement to obtain a current vehicle pose estimate $\poset{\timestep}$.
%
}
\end{figure}
The \visloc system consists of the following main components, presented at the beginning of this section. a)~We employ a state estimation framework for fusing wheel-odometry and visual localization constraints. b)~Our map-tracking module matches keypoints extracted from current camera images to landmarks from the map.
%
%
%
%
Furthermore, key information regarding our mapping pipeline is provided at the end of this section, and a schematic overview of \visloc can be found in Figure~\ref{fig:schema}.

\subsection{State Estimator}
\label{sec:methodology:subses:swe}
At the core of our localization system we employ a state estimation framework in information form, the dual representation of the (Extended) Kalman-Filter~\cite{burri2016generalized,strasdat2012visual}. 
Our state representation entails an estimate of the current transformation between the vehicle body coordinate frame $\bodyframe$ and the map reference frame $\mapframe$ for each time-step $\timestep$: $\state_\timestep \coloneqq \left[\poset{\timestep}\right]$.
Note that $\pose$ is an element of $\mathit{SE(3)}$, and thus represents all six degrees of freedom.
The corresponding rotations are represented by unit quaternions.
%
%
%
At every time-step $\timestep$, a set of $n$ simultaneously recorded camera images, and a relative odometry transformation measurement $\odometry$ are received.
A new state is created by forward-propagating the previous state estimate using the odometry measurement: $\poseguess{\timestep+1} \coloneqq \poset{\timestep} \odometry $.
It is used both in the map-tracking module as a pose prior, and as an initial linearization point in the filter update.

After localizing the current set of images, the states are updated by retrieving the \textit{MAP} estimate of the following cost function: 
\begin{align*}
\cost(\poset{\timestep+1}, \poset{\timestep}) &\coloneqq \parallel \fprior(\poset{\timestep}) \parallel^{2}_{\cov{\timestep}} \\ &+ \parallel \fodo(\poset{\timestep+1}, \poset{\timestep}) \parallel^{2}_{\covodo} \\ &+ \sum\limits_{i=1}^{m} \varphi(\floc(\poset{\timestep+1}, \poset{\timestep}))
\end{align*}
The prior pose and odometry factors, $\fodo$ and $\fprior$ respectively, follow a standard quadratic loss expression, while the localization re-projection factors $\floc$ employ a \textit{Huber} robust cost function $\varphi$ to account for possible wrong keypoint-landmark associations.
All factors follow a standard graph SLAM formulation, as described in~\cite{cadena2016past}.
We retrieve the \textit{MAP} estimate by iteratively minimizing the cost function~$\cost$ using the Levenberg-Marquardt in the GTSAM framework~\cite{dellaert2012factor}.

\subsection{Map-Tracking}
\label{sec:methodology:subsec:map_tracking}
\begin{figure}
\includegraphics[width=0.49\textwidth]{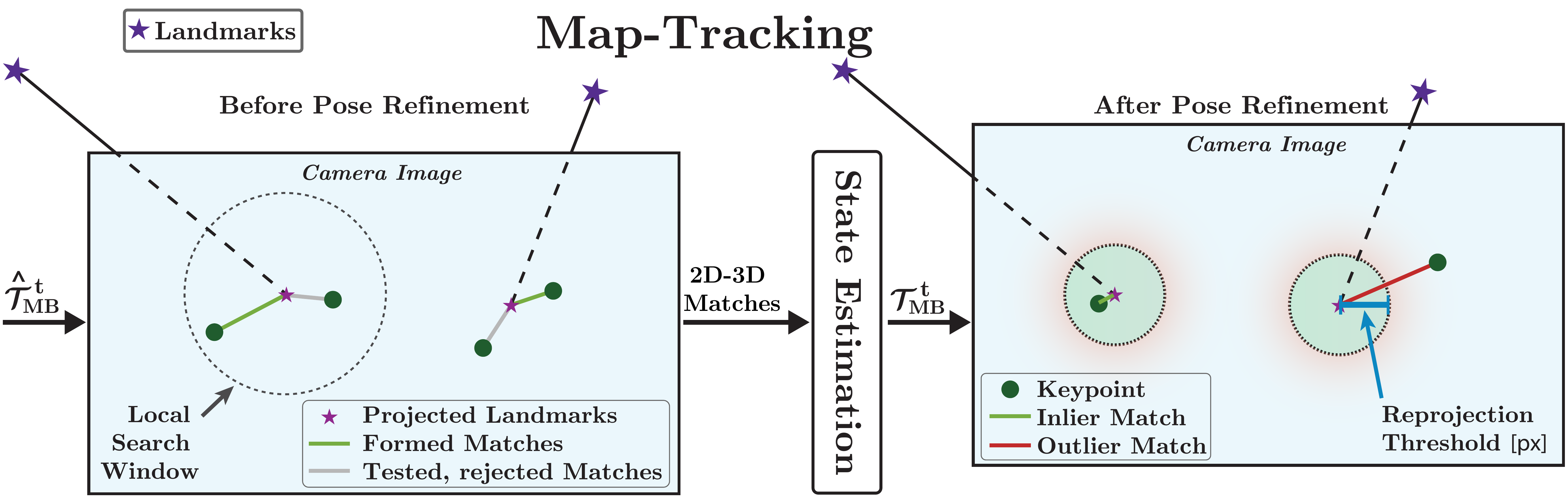}
\caption{\label{fig:schema_mt}
The map-tracking module extracts 2D features from current camera images, and matches them with 3D map landmarks locally in image space using a pose prior $\poseguess{\timestep}$.
The state estimation module fuses the visual \mbox{2D-3D} matches with the current wheel-odometry measurement to obtain a current vehicle pose estimate $\poset{\timestep}$.
%
}
\end{figure}
At every timestep $\timestep$, the forward-propagated pose $\poseguess{\timestep}$ represents a rough estimate of the vehicle's location at time $\timestep$ in the map.
With this, we can retrieve all landmarks from the map that have been observed from within a given distance around $\poseguess{\timestep}$.
%
%
%
Using $\poseguess{\timestep}$ and the extrinsics calibration between the vehicle body and the individual camera frames, the landmark $3D$ points are projected into the current set of images, and matched with extracted keypoints in the following way:
A landmark and a keypoint are only considered as a match candidate if their image space distance is smaller than $40px$.
This avoids forming geometrically inconsistent matches.
Further, a keypoint is preferably matched with the candidate landmark whose FREAK~\cite{Alahi2012FREAK} descriptor is closest to the FREAK descriptor of the keypoint, using the Hamming distance metric.
A \textit{descriptor distance threshold} $\ddt[px]$ is employed to limit the distance between the two descriptors, thus ensuring appearance consistency.
The resulting 2D-3D matches are fed back into the state estimator where they form the visual localization constraints for the state update at time $\timestep$.
After optimizing the vehicle pose $\poset{\timestep}$, the geometric consistency of every localization constraint is re-evaluated. 
For this, a \textit{reprojection threshold} $\rt$[px] is used to distinguish between inlier and outlier landmark observations.
While the localization factors of outlier observations are removed, the localization factors of inlier observations are marginalized out together with the previous pose $\poset{\timestep-1}$.
An illustration of the map-tracking algorithm can be found in Figure~\ref{fig:schema_mt}.
%
%
%
%

\subsection{Mapping}
\label{sec:methodology:subsec:mapping}
A base-map is built by tracking and triangulating local features along the trajectory of the first-session dataset.
The resulting $3D$ landmark points are added to the map together with their median feature descriptors.
Subsequently, more map sessions are added by localizing further datasets against the available (multi-)session map using map-tracking.
%
%
%
Note that all landmarks in the resulting multi-session map are expressed in one common frame of reference $\mapframe$.
Similar local localization and mapping algorithms have been used in our previous work~\cite{Muhlfellner2016SummaryMaps,Burki2016AppearanceBased}, to which we kindly refer the interested reader for more details.

\section{Evaluation}
\label{sec:evaluation}
This section presents evaluation results on the following three key aspects.
a)~In a parameter study, the optimal values for the most important parameters of our localization system are identified.
b)~We further investigate the influence of different binary descriptors on the localization performance.
c)~In long-term experiments across vastly different appearance conditions in outdoor environments, the localization accuracy and recall using map-tracking are evaluated, and compared with the accuracy and recall resulting from using a global localization algorithm.
%

The subsequent section first describes the \updrive vehicle platform, including the sensor set-up, computing infrastructure, and provides details on the online operation.
%
%
Additional sections are devoted to a brief description of the three dataset collections, and the evaluation metrics used in our experiments.

\subsection{The \updrive Platform}
\label{sec:eval:subsec:implemenation}
The \updrive vehicle is equipped with a surround-view camera system consisting of four cameras with fish-eye distorted lenses.
Images are recorded at $30Hz$ with a resolution of $640 \times 400$ pixels in gray-scale.
Furthermore, wheel tick encoders and a low-end IMU provide odometry measurements, which are fused with the visual localization constraints as described in Section \ref{sec:methodology:subsec:map_tracking}.
The vehicle and sample images from the camera system are depicted in Figure~\ref{fig:teaser}. 
Localization is run in real-time at $10Hz$ on a consumer-grade computer with an Intel i7 CPU and 16GB of RAM.
In particular, no GPU is required, neither for mapping, nor for localization.
%
%
Furthermore, for bootstrapping map-tracking, a position prior is generated with a consumer-grade GPS sensor, while the orientation hypothesis is generated from orientations of near-by map poses.
%
%
\subsection{Dataset Collections}
\label{sec:evaluation:subsec:datasets}
\subsubsection{UP-Drive}
%
%
The \updrive dataset collection consists of $32$ drives on the Volkswagen factory premises in Wolfsburg, Germany, recorded between December $2017$ and December $2018$.
The total driving distance is approximately $300km$.
The scenery resembles an urban environment, with busy streets, buses, zebra crossings, and pedestrians\footnote{
Sample images can found online at \url{https://github.com/ethz-asl/up-drive_visual_dataset/wiki/Sample-Images}}.
%
%
%
This dataset collection not only covers seasonal appearance changes and a wide range of different weather conditions, it also contains datasets recorded at dusk and night-time.
%
%
Five datasets, three from day-time, one at dusk, and one at night, are used to build a multi-session map.
%
%
The remaining $27$ datasets are used for evaluating the localization.
%
%
%

\subsubsection{\nclt}
The \ncltns~\cite{Carlevaris-Bianco2016} dataset collection consists of $27$ recordings collected with a Segway platform on the Michigan University campus between January $2012$ and April $2013$.
Analogous to the \updrive datasets, odometry poses based on wheel-tick encoders and an IMU sensor are fused in the state estimation framework.
A \textit{Ladybug 3} camera system is used, collecting images at $5Hz$ which are undistorted and down-scaled to a resolution of $808 \times 616$ pixels prior to being fed into our framework.
The visited routes vary considerably from dataset to dataset. 
However, there is an approximately $750m$ long outdoor segment that is traversed, with some minor deviations, in almost all datasets in either one or the opposite direction.
We therefore use this sub-segment of the campus for building a multi-session map using seven of the datasets.
The remainder of the datasets are used for evaluating the localization.
Similar to the \updrive datasets, the \nclt datasets cover seasonal and weather changes over an annual cycle.
%

\subsubsection{\kitti}
We further use $Sequence\text{ }00$ of the \kittins~\cite{geiger2013vision} visual odometry benchmark dataset in our evaluation.
It is the only \kitti dataset with significant segments of the trajectory revisited. 
We split the dataset in two parts, and use the first $170$ seconds for mapping, and the remainder for localization.
As opposed to the \updrive and the \nclt datasets, the appearance conditions in the \kitti drive thus remain similar between mapping and localization.
\subsection{Metrics}
\subsubsection{Localization Recall}
We measure localization recall~$\mathbf{r}$\footnotesize{[\%]}\normalsize~as the distance traveled while localized in relation to the total distance traveled in the respective dataset.
Localization at time $\timestep$ is deemed successful if there are at least $10$ inlier landmark observations after the pose optimization.
\subsubsection{Localization Accuracy}
The $6DoF$ localization accuracy is evaluated for each successfully localized set of images along the trajectory of a dataset by comparing the relative transformation between the estimated pose $\poset{\timestep}$ and the nearest vertex in the map, with the same quantity estimated by a reference solution~\cite{Burgard2009AComparison}.
For the \nclt datasets, ground-truth poses are available, which we employ to evaluate both the translation accuracy $\transerror$[m], and orientation accuracy $\orienterror$[deg].
Note that the availability of ground-truth poses is a unique feature of \ncltns, and the primary reason why we have decided to evaluate on the \nclt datasets, in addition to our self-collected \updrive datasets.

%
For the \updrive and \kitti datasets, no ground-truth poses are available.
Both dataset collections provide, however, poses estimated by an \textit{RTK GPS} sensor, which we use for producing a rough estimate of the localization accuracy on these datasets.
Since the \textit{RTK GPS} altitude estimates are unreliable, we only report on planar $\planarerror$[m] and lateral translation errors $\laterror$[deg] on the \updrive and \kitti datasets.
\subsection{Map-Tracking Parameter Study}
\label{sec:eval:subsec:sensitivity_and_convergence_analysis}
\begin{figure}
\includegraphics[width=0.49\textwidth]{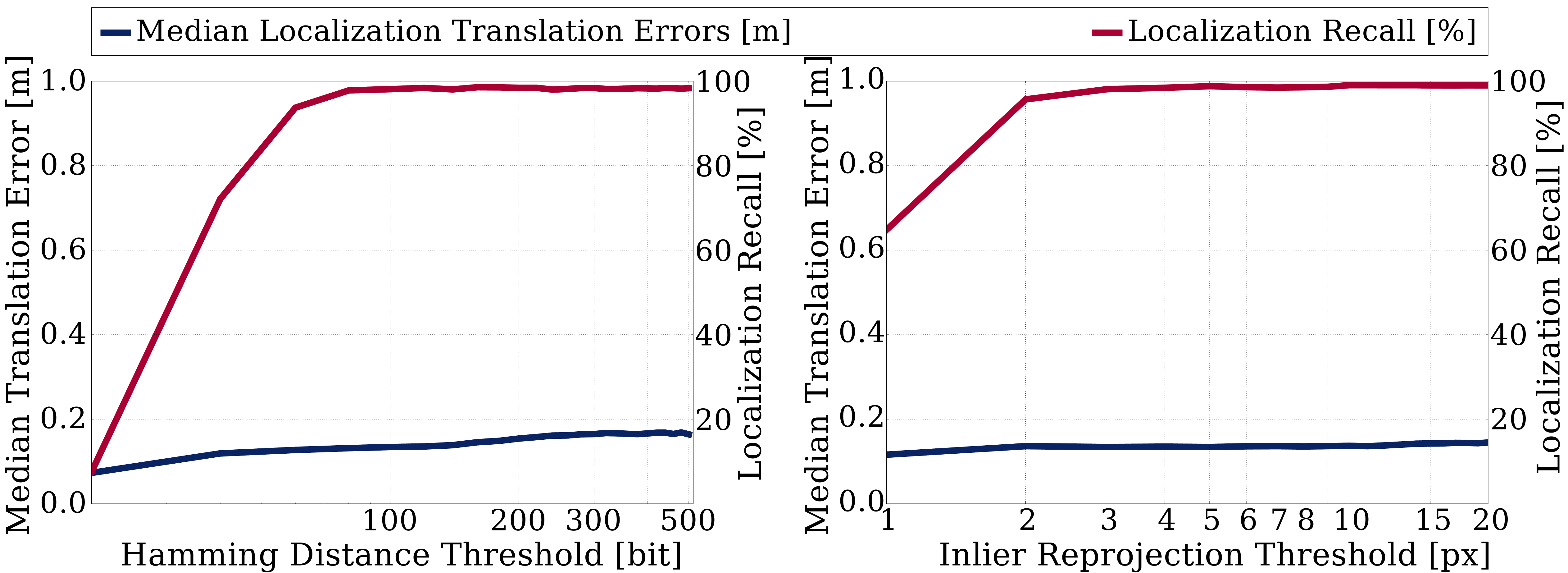}
\caption{\label{fig:nclt:2018-01-08:hamming_repro_sweep}
Localization recall $\recallmt$ (red) and median translation accuracy $\transmedianerror$ (blue) on the \nclt dataset from January $8^{th}$ 2012, in relation to increasing values of the \textit{descriptor distance threshold} $\ddt$ (left), and \textit{reprojection threshold} $\rt$ (right) respectively, on a logarithmic scale.
Even for very high values of $\ddt$ and $\rt$, the vehicle remains accurately localized.}
\vspace{-4mm}
\end{figure}
%
%
As described in Section~\ref{sec:methodology:subsec:map_tracking}, there are two main parameters guiding the formation of 2D-3D localization constraints in the map-tracking module, namely the \textit{descriptor distance threshold} $\ddt$[bits], and the \textit{reprojection threshold} $\rt$[px].
While the \textit{descriptor distance threshold} ensures appearance consistency by setting an upper bound on the descriptor distance for matching 2D keypoints with a 3D landmarks, the \textit{reprojection threshold} enforces geometric consistency by discarding localization constraints if their respective reprojection error after the pose update is more than $\rt$ pixels.

In Figure \ref{fig:nclt:2018-01-08:hamming_repro_sweep}, the localization recall and median localization error are shown for increasing values of $\ddt$, and $\rt$ respectively, for the \nclt \textit{2012-01-08} dataset.
A fixed value of $\ddt~=~100$bits, and $\rt~=~3$px is used unless the respective parameter is varied as indicated on the x-axis.
As expected, localization recall quickly rises with increasing $\ddt$ and $\rt$. 
Interestingly, the localization accuracy remains approximately constant, even for high values of $\ddt$ and $\rt$.
This may appear counter-intuitive at first.
A descriptor distance threshold greater than $25\%$ of the total descriptor length clearly allows for many wrong matches to be formed, and eventually ought to lead to false positive localizations.
In order to understand why this scenario does not occur, it is important to note that, as described in Section~\ref{sec:methodology:subsec:map_tracking}, the \textit{descriptor distance threshold} only serves to discard matches whose descriptor distance is above $\ddt$ bits.
If there are multiple match candidates for a given keypoint in the image, the matching algorithm still attempts to pick the landmark with the lowest descriptor distance. 
Therefore, as long as there \textit{are} sufficiently many correct matches that can be formed, our algorithm \textit{will} find them, even with a very lean \textit{descriptor distance threshold} $\ddt$. 

A similar effect exists for the \textit{reprojection threshold} too. 
As long as the pose prior is close to correct and there are sufficiently many valid 2D-3D matches, localization will not deviate from the correct trajectory, even with a very high $\rt$ threshold.

\begin{figure}
\includegraphics[width=0.49\textwidth]{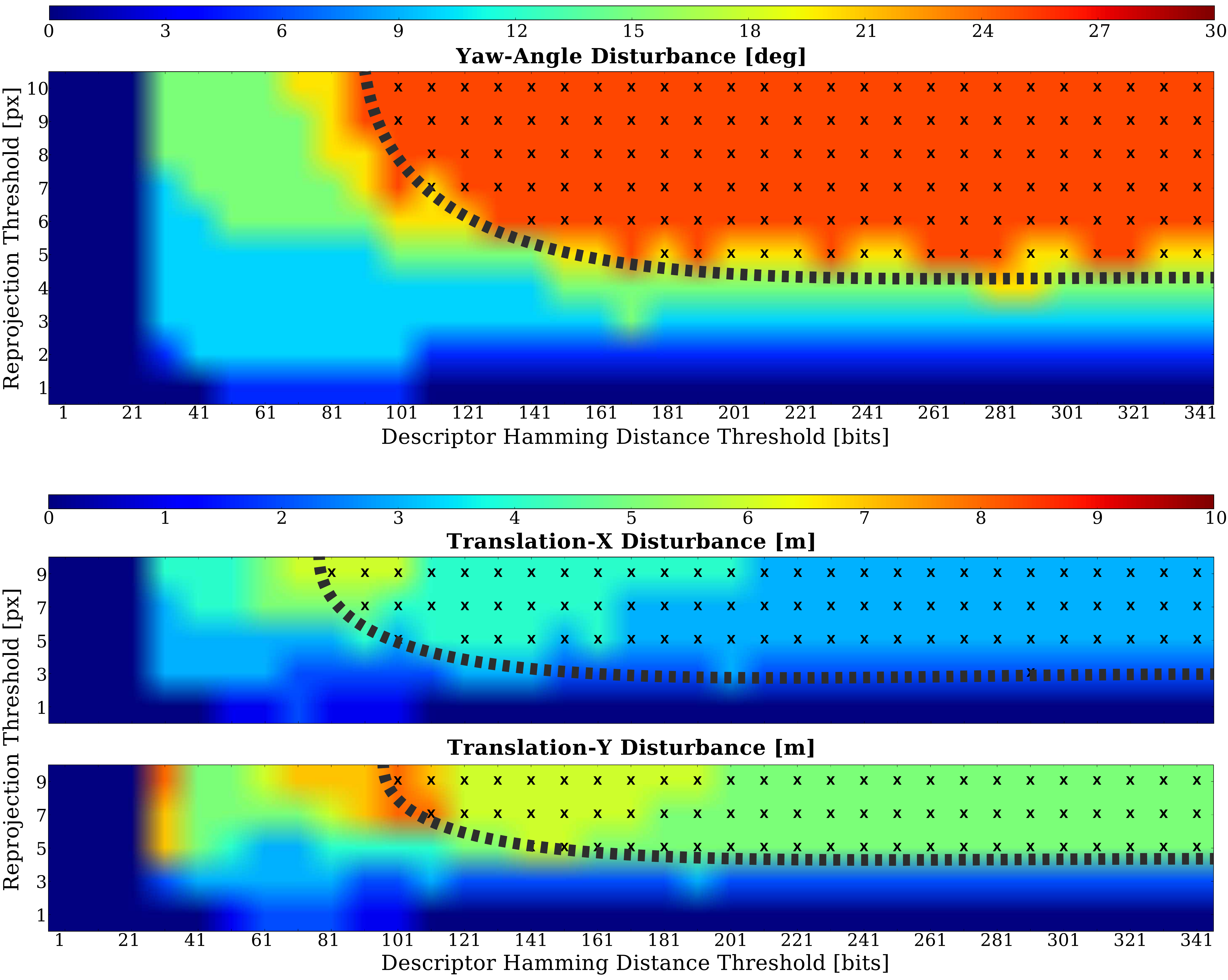}
\caption{\label{fig:convergence_basin}
Sensitivity of the \maptracking pose prior in relation to increasing values for the \textit{descriptor distance threshold}~$\ddt$, and \textit{reprojection threshold}~$\rt$.
The colors represent the maximum admissible degree of disturbance in yaw angle (top), longitudinal (middle), and lateral direction (bottom) leading to convergence of the localization to the true pose. 
The parameter combinations marked with `X' denote unsafe operating regions, where high prior pose disturbances lead to false-positive localizations.
In the remaining operating regions, localization fails if the prior pose disturbance is larger than the degree represented by the respective color.
All three modes of disturbances reveal a safe region for the choice of $\ddt$ and $\rt$ allowing for guaranteed convergence towards the correct pose, while tolerant to significant disturbance in the prior pose.}
\vspace{-4mm}
\end{figure}

Hence, as long as the vehicle is correctly localized, and there are enough valid localization matches \textit{possible}, our localization system will \textit{remain} correctly localized,

%
However, too high a value for $\ddt$ and $\rt$ may indeed derail the localization system if the pose prior is sufficiently wrong.
%
%
In the remainder of this section, we therefore aim at finding the range of values for $\ddt$ and $\rt$ that guarantee no false-positive localization, even if the pose prior is wrong.
Knowing this range is important in two ways.
Firstly, it defines a safe operating space for choosing $\ddt$ and $\rt$ where a positive localization feedback, such as a certain number of inlier landmark observations, can be trusted. 
Secondly, it reveals a maximum degree of pose prior disturbance that can be tolerated when bootstrapping the map-tracking algorithm with any kind of auxiliary global localization input such as consumer grade GPS, or a place-recognition module. 
In order to evaluate these properties, we have conducted a parameter sweep experiment, varying both values for $\ddt$ and $\rt$, as well as increasing the disturbance of the prior pose in yaw-angle, longitudinal, and lateral dimension separately. 
The resulting range of safe operating conditions is shown in Figure~\ref{fig:convergence_basin}. 
%
The colors indicate the maximum disturbance, before either bootstrapping map-tracking is no longer possible, or, marked with an `X', bootstrapping resulted in false-positive localization instead.
It can be seen that for all three modes for disturbance, there is a safe range for both $\ddt$, and $\rt$, that guarantee convergence to correct localizations, even for considerably inaccurate prior poses with up to $10$ degrees in yaw angle, and $3m$ meters in longitudinal and lateral direction.
Furthermore, taking the results from both Figure~\ref{fig:nclt:2018-01-08:hamming_repro_sweep}, and~\ref{fig:convergence_basin}, we find with $\ddt = 100bits$, and $\rt = 3.0px$, a safe choice of parameters yielding maximum recall and sufficient robustness for bootstrapping map-tracking with a consumer-grade GPS sensor.
%

\subsection{Binary Descriptor Comparison}
\label{sec:eval:subsec:descriptor_comparison}
\begin{figure}
\includegraphics[width=0.49\textwidth]{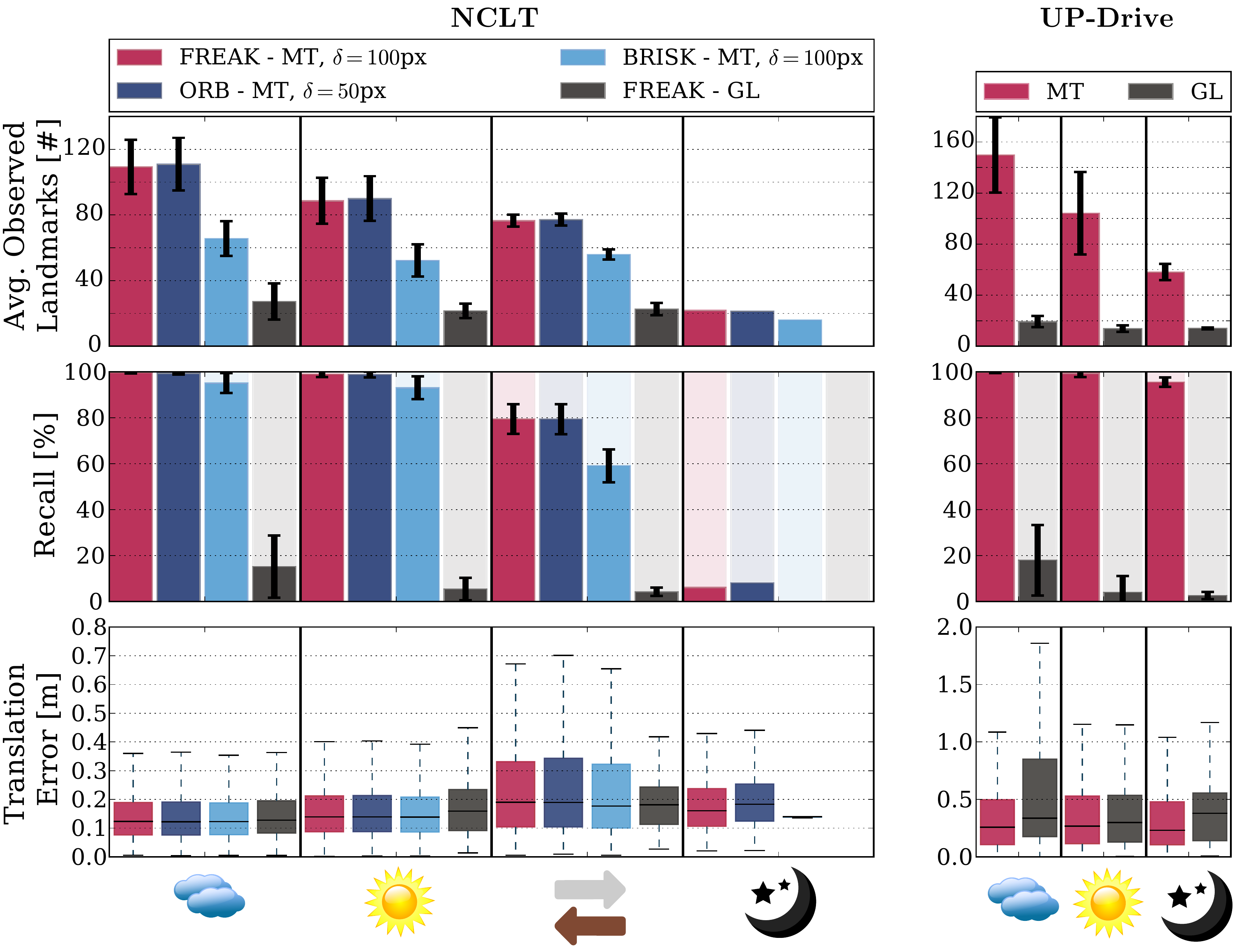}
\caption{\label{fig:nclt:descriptor_comparison}
Average number of observed landmarks (top), localization recall (middle), and translation accuracy on the \nclt datasets (left), and \updrive datasets (right).
For the \nclt datasets, the translation localization accuracy is evaluated using the \textit{ground-truth} poses, while for the \updrive datasets, the planar translation errors with respect to the \textit{RTK GPS} poses are shown.
The datasets are grouped into categories according to appearance conditions (cloudy or rainy, sunny, and night-time) and traversal direction (indicated by the two opposing arrows). 
The localization performance using \maptracking (MT) is compared with global localization (GL).
On the \nclt datasets, the \maptracking performance is further compared with different choices of binary descriptors.}
\vspace{-4mm}
\end{figure}
In addition to the \textit{descriptor distance threshold} and the \textit{reprojection threshold}, the type of descriptor is another pivotal design choice, as it influences not only the localization recall, but also the size of the map.
We restrict ourselves to the use of binary descriptors, as they can be matched very efficiently on a CPU-only platform, and compare the localization performance for three popular choices of binary local feature descriptors, namely FREAK~\cite{Alahi2012FREAK}, BRISK~\cite{leutenegger2011brisk}, and ORB~\cite{rublee2011orb}. 
Krajnik et al. have evaluated the influence of various local feature descriptors for visual teach-and-repeat in~\cite{krajnik2017image}.
They have, however, employed a global matching algorithm to find correspondences between two images recorded at the same location.
In contrast to that, our map-tracking algorithm employs a pose prior and performs a local search in the image space.
This variation in methodology leads to differing results as compared to~\cite{krajnik2017image}.
While the evaluation by Krajnik et al. suggests superior performance of BRISK as compared to FREAK and ORB, our experiments reveal worse performance of map-tracking with BRISK than with FREAK or ORB.
This emphasizes the strong influence of the specific feature matching algorithm with respect to the localization performance using different types of features.
%
%
In Table~\ref{table:descriptor_comparison}, the localization recall, the average number of observed landmarks, and the localization accuracy, are presented for different choices of descriptors, aggregated over all day-time \nclt datasets.
Note that the \textit{descriptor distance threshold} $\ddt$ is set at $100$bits for the two $64$ byte long descriptors FREAK and BRISK, and at $50$bit for the $32$ byte long ORB descriptors, thereby allowing the same relative fraction of bits to be different when forming localization matches in all three cases.
%
%
The performance using FREAK and ORB is nearly identical.
This is remarkable, as the descriptor size of the latter is only half of that of FREAK.
With BRISK, on the other hand, the average number of observed landmarks and the localization recall is significantly worse.
However, the impact on the localization accuracy is marginal, as only the pose estimates of successful localizations are considered.
%
%

A more detailed evaluation of the descriptor comparison can be found in Figure~\ref{fig:nclt:descriptor_comparison}, which shows the aforementioned metrics evaluated separately for groups of datasets formed according to the four categories exhibiting differing localization performance.
The category \textit{Cloudy} includes four, and the category \textit{Sunny} $12$, datasets labeled as (partially) cloudy, and sunny respectively, according to~\cite{Carlevaris-Bianco2016}.
The category \textit{Opposite} contains the two day-time datasets \textit{2012-11-04}, and \textit{2013-02-23} traversing the map in the opposite direction, while the \textit{Night} category represents the only night-time dataset from December $1^{st}$ $2012$.
The loss in recall with BRISK is primarily attributed to the two datasets traversing the map in opposite direction, where the recall with BRISK is approximately $20\%$ lower than with FREAK or ORB.
Contrary to that, the average number of observed landmarks remains roughly the same with BRISK across all three day-time categories, while FREAK and ORB observe significantly more landmarks when traversing the map in the predominant direction, both under cloudy skies, and in sunny conditions.

Based on these experiences, we suggest to use ORB as a binary descriptor for map-tracking, or FREAK in case there are no restrictions with respect to the map size.

\subsection{Localization Accuracy and Recall}
\label{sec:evaluation:subsec:localication_accuracy_and_recall}
\begin{table}[]
\begin{tabular}{r|lll}
\hline
\textbf{} & \textbf{FREAK} & \textbf{ORB} & \textbf{BRISK} \\ \hline
$\recallmt$\footnotesize{[\%]} &\textbf{96.89} +/- 6.62&\textbf{96.76} +/- 6.61&89.75 +/- 12.0\\
$\avginliers$[\footnotesize{\#}]&\textbf{92.97} +/- 49.94&\textbf{94.28} +/- 51.38&56.28 +/- 33.12\\
$\transmedianerror$[m]&0.14 [0.32]&0.14 [0.32]&0.14 [0.3]\\
\end{tabular}
\caption{\label{table:descriptor_comparison}
Descriptor comparison on the \nclt datasets, showing the average recall with \maptracking $\recallmt$\footnotesize{[\%]}, the average number of observed landmarks $\avginliers$\footnotesize{[\#]}, with standard deviations denoted by ``+/-'', and the median translation localization accuracy $\transmedianerror$[m], with the $90$~percentile denoted in square brackets.
%
%
}
\vspace{-10mm}
\end{table}
\begin{table*}
\center
\begin{tabular}{r|ll|lll}
\hline
 & $\recallmt$\footnotesize{[\%]} & $\recalllc$\footnotesize{[\%]} & \multicolumn{2}{l}{$\transmedianerror$ / \blue{$\planarmedianerror$, $\latmedianerror$}} & $\orientmedianerror$ \\ \hline \hline
NCLT&\textbf{96.89} +/-6.62&7.49 +/-8.59&\multicolumn{2}{c}{0.14 [0.32]}&1.23 [1.8]\\\hline
\hline
UP-Drive&\textbf{99.23} +/-1.75&8.94 +/-12.62&\blue{0.26 [0.88]}&\blue{0.12 [0.58]}&0.21 [0.33]\\
KITTI&\textbf{96.05}&94.24&\blue{0.43 [0.8]}&\blue{0.31 [0.62]}&0.26 [0.59]\\
\end{tabular}
\caption{\label{table:loc_performance_summary}
The aggregated localization performance on the \nclt, \updrive, and \kitti dataset(s), showing average localization recall with \maptracking $\recallmt$, and with \gloloc $\recalllc$, and the median translation ($\transmedianerror$) and orientation ($\orientmedianerror$) accuracy.
For \updrive and \kittins, \blue{planar~$\planarmedianerror$} and \blue{lateral~$\latmedianerror$} errors are shown instead of full $3DoF$ translation errors. 
Standard deviations are denoted by ``+/-'', and the $90$~percentile is shown in square brackets.
%
%
%
%
%
%
%
%
}
\vspace{-5mm}
\end{table*}
In order to fully rely on our visual localization system to control the car in the \updrive project, a high localization recall with an accuracy below $0.5m$ is paramount, as only short driving segments with no localization may be bridged with wheel-odometry before the car may deviate from its designated lane. 
%
%
We compare the localization recall and accuracy of our localization system using map-tracking with the metric global localization algorithm based on the work presented in~\cite{lynen2015get} and available in the \textit{maplab} framework~\cite{schneider2018maplab}.
We refer to the results using this algorithm with $GL$ in the respective figures and tables.
Both algorithms, that is map-tracking and global localization, operate on the same multi-session maps, using the same landmarks.
Note, however, that the global localization algorithm is fundamentally different to the map-tracking module presented in this paper, as in contrast to the former, the latter is able to exploit a pose prior.
By including this comparison, we aim at highlighting the gain in localization recall attainable by using a local localization algorithm, as opposed to relying only a global localization algorithm.
Map-tracking does, however, require \textit{some} global localization module for bootstrapping, or re-localizations.
As described in Section~\ref{sec:eval:subsec:implemenation}, a consumer grade GPS sensor serves this role on the \updrive vehicles.

The localization recall with map-tracking $\recallmt$\footnotesize{[\%]}\normalsize, and with global localization $\recalllc$\footnotesize{[\%]}\normalsize, and the localization accuracy are shown in Table~\ref{table:loc_performance_summary}, aggregated over all datasets of the three collections.
Note that the \nclt night-time dataset from December $1^{st}$ is excluded in the table.
%
While \maptracking attains close to $100\%$ recall on all three dataset collections, global localization fails for extended periods on the \nclt and \updrive datasets which both exhibit pronounced appearance change.
On the \kitti drive, however, the appearance condition only undergo minor change, and global localization achieves with $94\%$ a similarly high recall as map-tracking.
%
%
This illustrates the challenge in finding enough correct feature matches with a global localization algorithm in multi-session maps that cover outdoor environments with various different appearance conditions.
Solely relying on a global localization algorithm in these environments may thus not be sufficient to guarantee reliable localization in real-world applications.
As our results show, exploiting a pose prior can help to significantly increase the reliability of the localization.

We further note that the planar median localization accuracy in \updrive and \kitti are below $0.5m$.
Note that due to the different kind of reference sensors, these numbers are not directly comparable with the localization accuracy attained on the \nclt datasets, with the latter exhibiting a median translational accuracy of $11cm$. 
Furthermore, the \kitti vehicle is equipped with only a forward facing camera, while the \updrive vehicle has a surround view camera rig.
This results in less strictly constraint position estimates on the \kitti dataset, which translate into significantly lower planar and lateral localization accuracy.
For the driving performance, the lateral errors are most important. 
On the \updrive datasets, the median lateral error is below $15cm$, which is sufficient for a smooth steering of the car.

The median orientation errors are less effected by the difference in the camera rigs, and are well below $0.5$ degrees for both the \updrive and \kitti datasets.
In contrast to that, the orientation errors on the \nclt datasets are higher due to more vivid roll and pitch motions of the Segway platform, as compared to the car platforms in case of \updrive and \kittins.

A more detailed analysis of the localization recall and planar accuracy on the \nclt and \updrive datasets is shown in Figure~\ref{fig:nclt:descriptor_comparison}, with datasets grouped into categories as described in Section~\ref{sec:eval:subsec:descriptor_comparison}.
There are $10$ drives of the \updrive dataset collection in the \textit{Cloudy}, $15$ in the \textit{Sunny}, and two in the \textit{Night} category respectively.
Recordings in rainy conditions are categorized as \textit{Cloudy}, since there is little difference in performance on rainy datasets as opposed to in dry cloudy conditions.
Map-tracking reaches virtually $100\%$ recall with a median localization accuracy of around $10cm$ for all the \nclt day-time datasets that traverse the map in the primary direction.
The same high recall is also achieved for all day-time \updrive datasets, with a planar median localization accuracy with respect to RTK GPS of approximately $20cm$.
The additional challenge for visual localization in sunny conditions is, however, reflected in a lower average number of observed landmarks in case of map-tracking, and in a significantly worse recall using global localization.
Recall using map-tracking remains, however, unaffected.
%
%
%

In contrast to that, map-tracking performs significantly worse on the two \nclt datasets that traverse the map in the opposite direction, with considerably lower recall, and slightly lower localization accuracy. %
This is understandable, given that there is only one map session in opposite direction, whereas there are six traversing the map in the primary direction.
However, this also reveals the limitations of matching landmarks projected into the cameras field-of-views under considerable viewpoint change.
Here it is important to note the asymmetry of the \textit{Ladybug} camera rig when traversing in the opposite direction, as the surround view is covered by an odd number of five cameras.

\begin{figure}
\includegraphics[width=0.49\textwidth]{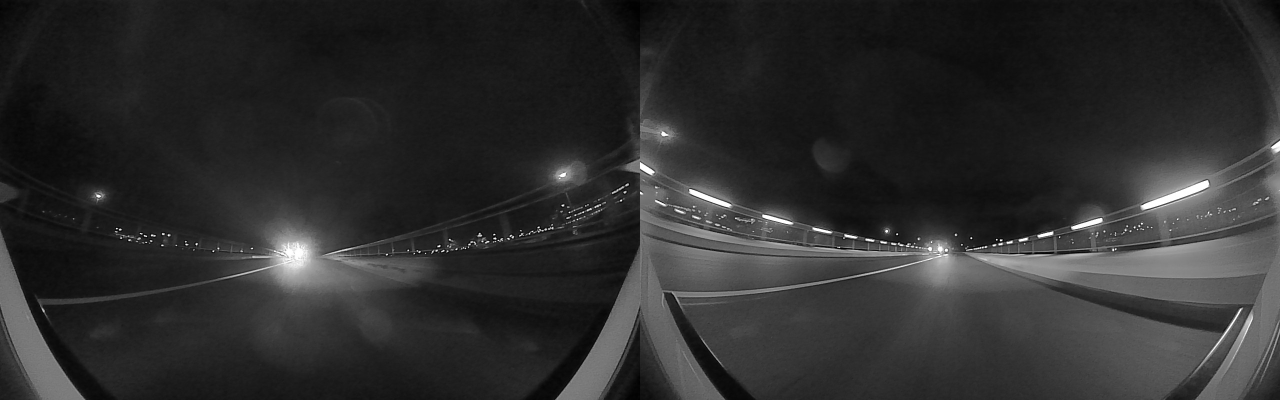}
\caption{\label{fig:fail_loc_sample_images}
On the left, a sample image of a trajectory segment that fails to localize at night.
A lack of structure and street lighting renders it unfeasible to match a sufficient number of map landmarks.
A few meters later, street lighting is present (right side), and localization is picked up again.}
\vspace{-4mm}
\end{figure}

Furthermore, the only \nclt night-time dataset from December $1^{st}$ $2012$ fails to localize along most parts of the trajectory.
Not only is this the only available recording under night-time conditions, but the Segway also traverses the map in the opposite direction, further exacerbating localization.
A lack of more recordings from dusk or night-time renders it impossible to augment the multi-session map with the appearance conditions at night-time, and thus the conditions in this dataset lie outside the appearance coverage of the map.
In contrast to the \nclt datasets, the \updrive datasets contain multiple recordings under both dusk and night-time conditions, allowing to extend the appearance coverage of the multi-session map with these conditions.
Therefore, localization at night is successful in this case, even though the respective average recall is slightly less than $100\%$ for the \updrive night-time datasets.
%
This minor drop in recall is mainly attributed to the night-time recording from December $11^{th}$, which only attains a recall of $92\%$.
%
%
Sample images of the route segment where localization fails on this dataset are depicted in Figure~\ref{fig:fail_loc_sample_images}.
In this part of the route, the car is driving up North on a ramp crossing numerous rail tracks. 
With a lack of both street lamps and near-by building structures, there are hardly any stable visual cues in this section, and our localization system fails to match sufficiently many landmarks from the map.
Only later along the ramp, once artificial lighting on the railing to the left and right of the road boundary is present, localization is picked up again.
This example demonstrates the current limitations of visual localization in night-time conditions. 
Even with high-performance CMOS cameras providing remarkably bright images at night, a certain amount of artificial street lighting and human made structure in the vicinity is required.

%
%
%
%
%
%

\section{conclusions}
\label{sec:conclusions}
This paper presented a reliable visual localization system for urban outdoor environments.
An extensive evaluation on several hundreds of kilometers of real-world driving conditions over the course of more than a year has demonstrated that our localization system is able to meet the requirement of high localization recall at high accuracy. 
Thereby, the appearance conditions encountered in our experiments not only cover various challenging weather conditions, wet road surfaces, sun reflections, and seasonal changes, but also night-time conditions.
A comparison with a state-of-the-art global metric localization algorithm has revealed a large increase in recall attainable by instead employing a local localization algorithm, such as the map-tracking algorithm described in this paper.
Additionally, a comparison of binary feature descriptors suggests superior performance of map-tracking when using FREAK or ORB, as compared to using BRISK.
In a thorough parameter study, we have further investigated the boundary conditions of our map-tracking module and validated a safe range for selecting the most critical parameters in order to guarantees reliable localization.
%
%

\addtolength{\textheight}{-10cm}   





\section*{ACKNOWLEDGMENT}

This project has received funding from the EU H2020 research project under grant agreement No 688652 and from the Swiss State Secretariat for Education, Research and Innovation (SERI) under contract number 15.0284.



{\small
\bibliographystyle{IEEEtran}
\bibliography{mendeley_2}
}

\end{document}